\documentclass[letterpaper, 10 pt, journal, twoside]{IEEEtran}
\usepackage{amsmath,amsfonts}
\usepackage{algorithmic}
\usepackage{algorithm}
\usepackage{array}
\usepackage[caption=false,font=normalsize,labelfont=sf,textfont=sf]{subfig}
\usepackage{hyperref}
\usepackage{todonotes}
\usepackage{textcomp}
\usepackage{stfloats}
\usepackage{url}
\usepackage{verbatim}
\usepackage{graphicx}
\usepackage{balance}
\usepackage{cite}
\usepackage{amsmath,amsfonts}
\usepackage{threeparttable}
\usepackage{multirow}
\usepackage{booktabs}  
\newcommand{\compacteqspacing}{%
  \setlength{\abovedisplayskip}{2pt plus 1pt minus 1pt}%
  \setlength{\abovedisplayshortskip}{2pt plus 1pt minus 1pt}%
  \setlength{\belowdisplayskip}{2pt plus 1pt minus 1pt}%
  \setlength{\belowdisplayshortskip}{2pt plus 1pt minus 1pt}%
}

\begin{document}

\title{MANGO-Grasp: Mahalanobis Fields over Geometry-Oriented 3D Gaussians for Cross-Embodiment Dexterous Grasping}

\author{
Heng Zhang$^{1,2}$~\IEEEmembership{Graduate Student Member,~IEEE},
Kevin Yuchen Ma$^{1,3}$~\IEEEmembership{Graduate Student Member,~IEEE}, \\
Mike Zheng Shou$^{3}$~\IEEEmembership{Senior Member,~IEEE},
Weisi Lin$^{2*}$~\IEEEmembership{Fellow,~IEEE} and
Yan Wu$^{1*}$~\IEEEmembership{Senior Member,~IEEE}

\thanks{Manuscript received July 11, 2026; revised October 12, 2026; accepted November 16, 2026.}

\thanks{This paper was recommended for publication by Editor Markus Vincze upon evaluation of the Associate Editor and Reviewers' comments.}

\thanks{*Corresponding authors: Weisi Lin and Yan Wu.}

\thanks{$^{1}$Heng Zhang, Kevin Yuchen Ma, and Yan Wu are with the Embodied AI and Robotics Chapter, Institute of Advanced Intelligence and Computing (IAIC), Agency for Science, Technology and Research (A*STAR), Singapore
{\tt\footnotesize wuy@i2r.a-star.edu.sg}.}

\thanks{$^{2}$Heng Zhang and Weisi Lin are with the College of Computing and Data Science, Nanyang Technological University, Singapore
{\tt\footnotesize HENG018@e.ntu.edu.sg, wslin@ntu.edu.sg}.}

\thanks{$^{3}$Kevin Yuchen Ma and Mike Zheng Shou are with Show Lab, National University of Singapore, Singapore
{\tt\footnotesize yuchen\_ma@u.nus.edu, mikeshou@nus.edu.sg}.}

\thanks{Digital Object Identifier (DOI): see top of this page.}
}

\markboth{IEEE Robotics and Automation Letters. Preprint Version. Accepted November, 2026}
{Zhang \MakeLowercase{\textit{et al.}}: MANGO-Grasp: Mahalanobis Fields over Geometry-Oriented 3D Gaussians for Cross-Embodiment Dexterous Grasping} 

\maketitle

\begin{abstract}
Cross-embodiment dexterous grasping aims to synthesize stable grasps
across heterogeneous multi-fingered hands with little or no embodiment-specific
tuning. Existing interaction-centric methods achieve promising results, but their
object representations often underrepresent local surface geometry, hand
descriptors do not explicitly capture both morphology and kinematics, and
hand--object interaction modeling is typically isotropic.
We propose MANGO-Grasp, an anisotropic interaction framework that
represents objects as geometry-oriented 3D Gaussian primitives to better
preserve local surface geometry, and encodes hands into pretrained descriptors
that capture both morphology and kinematics.
Hand--object interaction is modeled using anisotropic Mahalanobis fields
that reflect the directional nature of contact and guide a shared
optimization for grasp realization across embodiments.
On the CMAP and MultiGripperGrasp benchmarks, MANGO-Grasp outperforms the strongest seen-hand baseline by up to \textbf{8.24} percentage points in simulation. It also transfers zero-shot to the unseen SharpaWave hand, improving over the strongest zero-shot baseline by up to \textbf{16.57} percentage points, and achieves \textbf{86\%} success in real-world experiments.
The code and additional materials are available at \url{https://connor-zh.github.io/MANGO-Grasp/}.
\end{abstract}

\section{Introduction}
Dexterous grasping is a fundamental capability for robotic manipulation, yet synthesizing stable grasps remains challenging due to the high-dimensional
kinematics, complex contact constraints, and morphology-dependent feasibility of multi-fingered hands. Existing pipelines often rely on hand-engineered, embodiment-specific optimization~\cite{liu2021synthesizing, wang2022dexgraspnet,
10160314} or train dedicated models for individual hands~\cite{xu2024dexterous,
weng2024dexdiffusergeneratingdexterousgrasps,
zhong2025dexgraspanythinguniversalrobotic}. As robotic hardware continues to diversify, such cost scales poorly, motivating \emph{cross-embodiment dexterous grasping}: unified models that can generalize across different hands.

\begin{figure}[t]
    \centering
    \includegraphics[
        width=0.85\linewidth,
    ]{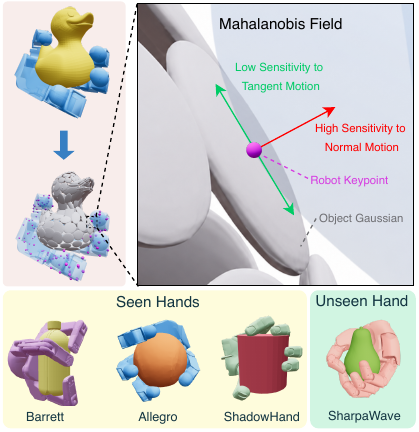}
    \vspace{-10pt}
    \caption{Overview of the proposed anisotropic interaction formulation for cross-embodiment dexterous grasping. By representing objects with geometry-oriented 3D Gaussian primitives and hands with keypoints, we use Mahalanobis fields to model keypoint--primitive interaction with
    low sensitivity to surface tangential motion and high sensitivity to surface normal
    motion. By learning such transferable interaction fields, our framework generalizes across seen hands and transfers
    zero-shot to an unseen hand.
    }
    \vspace{-12pt}
    \label{fig:teaser}
\end{figure}

Recent progress in cross-embodiment grasping has moved from hand- or
object-centric formulations toward interaction-centric modeling. 
Hand-centric
methods directly regress embodiment-specific configurations from object
observations~\cite{xu2023unidexgrasp, wan2023unidexgrasp++,
wang2025unigrasptransformer,
zhang2026machagraspmorphologyawarecrossembodimentdexterous}, but often struggle
to transfer across hands with substantially different kinematics. Object-centric
methods instead predict hand-agnostic contact targets~\cite{8972562,
li2023gendexgrasp, attarian2023geometry, c++}, but leave a realization gap
between object-side contacts and feasible hand configurations.
Recent
\emph{interaction-centric} methods have achieved promising results by modeling
hand--object spatial relations~\cite{11127754DRO,
fei2025trograspefficientgraph}, retaining embodiment awareness while capturing
transferable contact structure. However, existing interaction-centric methods remain limited by how object geometry, hand morphology and kinematics, and hand--object interaction are represented.

On the object side, prior object representations commonly rely on
uniformly sampled point clouds~\cite{11127754DRO} or point-cloud
patches~\cite{fei2025trograspefficientgraph}. 
Such representations do not adapt sampling density to geometric complexity, which may lead to insufficient sampling in high-curvature regions and redundant sampling in smooth regions~\cite{alexandros2021revisiting}.
Moreover, they do not explicitly encode local surface geometry. 
Consequently, hand--object interaction is modeled as isotropic
Euclidean proximity,
while contact is direction-dependent: tangential sliding and normal separation have distinct physical meanings.

On the robot side, existing representations either encode URDF-derived link
geometry and spatial pose~\cite{fei2025trograspefficientgraph}, or learn dense
configuration-invariant keypoint correspondences~\cite{11127754DRO}. These designs
improve embodiment awareness, but they do not explicitly couple morphological
identity with configuration-dependent motion. Since grasp feasibility depends on
both what the shape of a hand part is and how it can move, transferable robot descriptors
should jointly encode morphology and kinematic behavior.

Motivated by these observations, we propose an \emph{anisotropic interaction formulation} for cross-embodiment dexterous grasping, as shown in Fig.~\ref{fig:teaser}. On the object side, we adapt 3D Gaussian Splatting~\cite{kerbl20233d}
into a geometry-oriented primitive generation pipeline that produces a
fixed-size set of \emph{surface-aligned, plate-like 3D Gaussian primitives}
with assigned outward normals, preserving local tangent-normal geometry while adaptively allocating the fixed primitive budget according to local geometric complexity through geometry-driven densification and curvature-aware fixed-budget selection (Sec.~\ref{sec:gaussian}).
On the robot side, we pretrain morpho-kinematic keypoint descriptors by combining morphology-identity learning~\cite{11127754DRO} with a kinematic-awareness objective that encourages the descriptors to capture how each keypoint's position changes with the hand configuration.

Given these object and robot representations, we formulate robot--object
interaction as Mahalanobis fields over hand keypoints and object Gaussian primitives.
Unlike isotropic Euclidean distance, these fields measure
keypoint--primitive proximity using each primitive's surface
orientation and anisotropic scales, weighting normal displacement
more strongly than tangential displacement.
Through the anisotropic scales of each primitive, broad planar primitives induce larger
lateral tolerance, whereas compact primitives impose stricter localization. The
predicted Mahalanobis fields serve as both interaction prediction targets and surface-aware
guidance signals for grasp realization, which is performed
using a shared optimization setup across embodiments.

We evaluate our method on CMAP~\cite{li2023gendexgrasp} and
MultiGripperGrasp~\cite{casas2024multigrippergrasp} with three seen hands and one unseen hand. Our method achieves seen-hand simulation success rates of $97.59\%$ and $89.47\%$. The same model transfers zero-shot to the unseen SharpaWave hand,
obtaining $84.17\%$ and $81.47\%$ success on the two benchmarks, and reaches
$86\%$ success in the real world. Ablation studies
further validate the contribution of each proposed component.

Our main contributions are as follows.
\begin{itemize}
    \item We adapt a 3D Gaussian Splatting pipeline to represent object geometry as a fixed-size set of geometry-oriented 3D Gaussian primitives.

    \item We propose a \emph{morpho-kinematic} robot pretraining paradigm that learns hand-keypoint descriptors by capturing morphology identity and kinematic awareness.
    
    \item We formulate cross-embodiment hand--object interaction as \emph{
    Mahalanobis fields} between hand keypoints and object Gaussian primitives, enabling anisotropic interaction modeling and surface-aware grasp realization.

    \item We validate the method in simulation across seen and unseen
    embodiments and on a real unseen SharpaWave hand, demonstrating strong
    benchmark performance, zero-shot transfer, and direct hardware deployment.

\end{itemize}

\section{Related Work}
In this section, we review prior work on robot encoding, object encoding, and interaction modeling.

\noindent\textbf{Robot Encoding.}
Object-centric methods encode the robot implicitly through hand-agnostic
contacts or geometry embeddings~\cite{8972562,li2023gendexgrasp,attarian2023geometry,
wu2025cedexcrossembodimentdexterousgrasp}, but leave a gap between contact
prediction and grasp realization. Hand-centric methods expose
URDF-derived graphs, link features, kinematic structures, or eigengrasp bases~\cite{fei2025trograspefficientgraph,
wu2026unimorphgraspdiffusionmodelmorphologyawareness,
zhang2026machagraspmorphologyawarecrossembodimentdexterous}, yet these inputs
do not guarantee that the learned features capture morphology and kinematics.
DRO~\cite{11127754DRO} pretrains configuration-invariant keypoint descriptors
through contrastive alignment, capturing morphology-level correspondence but
not explicitly supervising kinematic behavior. We build on this paradigm by
adding kinematic-awareness supervision, producing robot descriptors
that encode both hand morphology and kinematics.

\noindent\textbf{Object Encoding.} Most cross-embodiment methods represent objects as point clouds~\cite{8972562,11127754DRO,
wu2025cedexcrossembodimentdexterousgrasp,
wu2026unimorphgraspdiffusionmodelmorphologyawareness,
yuan2025crossembodiment,
zhang2026machagraspmorphologyawarecrossembodimentdexterous}
or point-sampled patches and graphs~\cite{attarian2023geometry,c++,fei2025trograspefficientgraph}.
However, these representations allocate capacity through sampling
density and lack explicit surface geometry modeling. 
Inspired by Gaussian-based visual reconstruction~\cite{kerbl20233d,guédon2023sugarsurfacealignedgaussiansplatting,huang20242d},
we encode objects as a fixed-size set of geometry-oriented 3D Gaussians,
preserving local surface geometry while adapting primitive density
to geometric complexity.

\noindent\textbf{Interaction Modeling.} Interaction-centric methods achieve promising results by modeling spatial relations between hands and objects. DRO~\cite{11127754DRO}
represents interaction as a Euclidean distance matrix between robot
keypoints and object points, while TRO~\cite{fei2025trograspefficientgraph}
models transformations between hand links and object patches. However, their interaction metrics are limited in directional expressiveness. Euclidean distances treat all directions equally,
and patch-level transformations do not encode the local
tangent-normal asymmetry of contact. We instead define interaction as
anisotropic fields between robot keypoints and object Gaussian primitives.

\section{Method}

\subsection{Problem Formulation}

We address \emph{cross-embodiment dexterous grasp synthesis}: given
a target object and a robotic hand embodiment $e$ from a set of
heterogeneous candidates, produce a hand configuration $\mathbf{q}^*$
(joint angles together with a 6-DoF wrist pose) that yields a
kinematically feasible and stable grasp.

\noindent\textbf{Object Input.}
Each object mesh is processed into $G$ 3D Gaussian primitives. We set $G=256$ empirically to balance geometry coverage and downstream interaction matrix size.
{
\compacteqspacing
\begin{equation}
\mathcal{G}=\{g_j\}_{j=1}^{G}, \quad
g_j = (\boldsymbol{\mu}_j, \mathbf{R}_j, \boldsymbol{\sigma}_j, \mathbf{n}_j),
\end{equation}
}
where $\boldsymbol{\mu}_j\in\mathbb{R}^3$ denotes the primitive center,
$\mathbf{R}_j\in SO(3)$ its orientation,
$\boldsymbol{\sigma}_j\in\mathbb{R}^3_{+}$ its anisotropic scale, and
$\mathbf{n}_j\in\mathbb{S}^2$ its outward surface normal.

\noindent\textbf{Robot Input.}
Following~\cite{11127754DRO}, we represent each embodiment $e$ by $N$ surface keypoints, obtained once via farthest point sampling over the link meshes at a canonical pose. We set $N=256$ empirically to balance hand-surface resolution and the computational cost of downstream interaction matrix. Let $\mathcal{P}_e(\mathbf{q})$ denote the resulting keypoint set of embodiment $e$ at configuration $\mathbf{q}$, obtained by mapping the canonical keypoints through analytical forward kinematics:
{
\compacteqspacing
\begin{equation}
\mathcal{P}_e(q)=\{\mathbf{p}_i(q)\}_{i=1}^{N}\subset\mathbb{R}^3,\end{equation}
}
where $\mathbf{p}_i$ is the $i$-th base-frame keypoint, corresponding to a fixed surface site across configurations.

\noindent\textbf{Interaction Fields Generation and Grasp Realization.}
Our method first generates hand--object interaction fields
and then uses these fields to guide optimization of the
hand configuration.
Given hand keypoints $\mathbf{P}^{0}_e=\mathcal{P}_e(\mathbf{q}^{0})$
from an initial configuration $\mathbf{q}^{0}$ and object primitives
$\mathcal{G}$, a model $f_\theta$ generates target Mahalanobis fields,
represented by a matrix $\hat{\mathbf{M}}\in\mathbb{R}_+^{N\times G}$.
Each entry $\hat{M}_{ij}$ specifies the target Mahalanobis distance between
keypoint $i$ and primitive $g_j$ at the target grasp. The final grasp is
realized by solving:
{
\compacteqspacing
\begin{equation}
\mathbf{q}^*
=
\arg\min_{\mathbf{q}\in\mathcal{Q}_e}
\mathcal{L}_{\mathrm{rec}}
\big(\mathbf{q};\hat{\mathbf{M}},\mathcal{G}\big),
\end{equation}}
initialized from $\mathbf{q}^{0}$. 
$\mathcal{Q}_{e}
=\{\mathbf{q}\mid
\underline{\mathbf{q}}_{r}\leq\mathbf{q}_{r}
\leq\overline{\mathbf{q}}_{r}\}$
enforces revolute joint limits.
$\mathbf{q}_{r}$ denotes revolute-joint angles, with $\underline{\mathbf{q}}_{r}$ and $\overline{\mathbf{q}}_{r}$ as lower and upper limits.
The grasp-realization objective $\mathcal{L}_{\mathrm{rec}}$
combines Mahalanobis field guidance with penetration and
self-collision energies (see Eq.~\eqref{eq:optimization}).
\subsection{Method Overview}
\begin{figure*}[t!]
    \centering
    \includegraphics[
        width=0.95\textwidth,
    ]{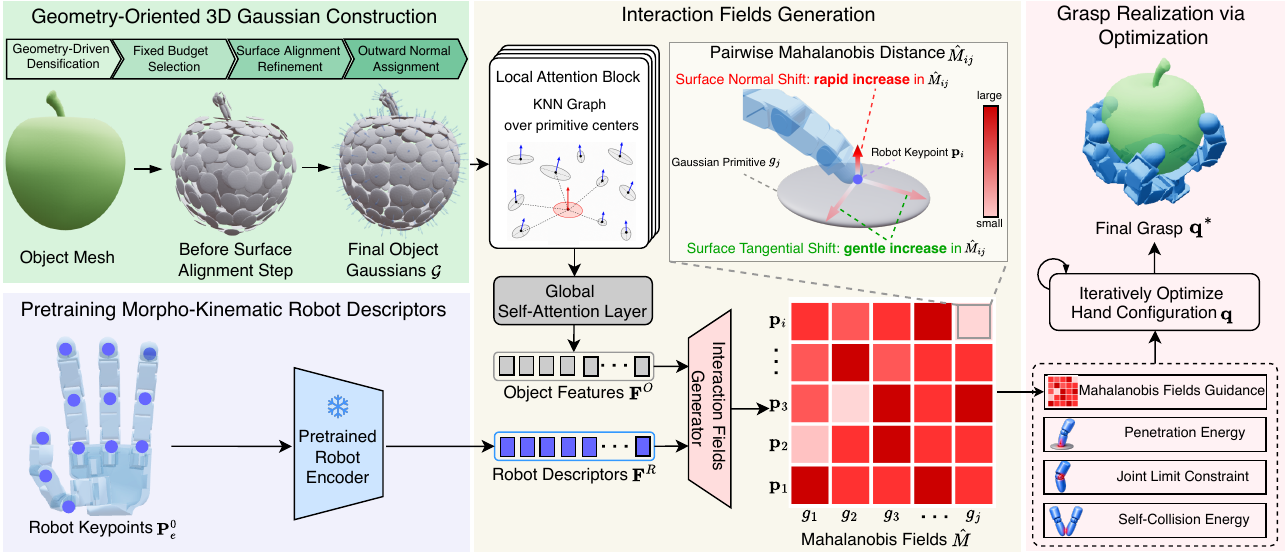}
    \vspace{-10pt}
    \caption{
    Method overview. The object mesh is converted into a fixed-budget set of surface-aligned, plate-like 3D Gaussian primitives $\mathcal{G}$ with outward normals. The
    primitives are then encoded by a stack of local attention blocks, operating
    on a KNN graph built over primitive centers, followed by a global
    self-attention layer, yielding Object Features $\mathbf{F}^O$. In parallel, initial robot keypoints $\mathbf{P}^{0}_e$ are encoded by a pretrained robot encoder into morpho-kinematic Robot Descriptors $\mathbf{F}^R$. The Interaction Fields Generator fuses both streams to generate the Mahalanobis fields $\hat{\mathbf{M}}$,
    which encode the interaction between each robot keypoint and each object
    primitive: the field value rises rapidly along the
    surface-normal direction and varies gently within the local tangent plane. The predicted fields then guide optimization from $\mathbf{q}^0$ to the final grasp $\mathbf{q}^*$ under joint-limit constraints, penetration energy, and self-collision energy.
    }
    \vspace{-15pt}
\label{fig:architecture}
\end{figure*}
As shown in Fig.~\ref{fig:architecture}, MANGO-Grasp consists of four components: Geometry-Oriented 3D Gaussian
Construction (Sec.~\ref{sec:gaussian}), Pretraining Morpho-Kinematic Robot Descriptors (Sec.~\ref{sec:pretraining}), Interaction Fields Generation
(Sec.~\ref{sec:prediction}), and Grasp Realization via Optimization
(Sec.~\ref{sec:recovery}).
\subsection{Geometry-Oriented 3D Gaussian Construction}
\label{sec:gaussian}
We adapt 3D Gaussian Splatting~\cite{kerbl20233d}, implemented with
Nerfstudio~\cite{nerfstudio}, to convert an object mesh into a fixed-budget
set of surface-aligned, plate-like 3D Gaussian primitives with assigned outward normals that encode
geometry rather than appearance. The pipeline has four stages:
geometry-driven densification allocates primitives according
to local geometric complexity using mesh-derived signals; fixed-budget selection retains $G$ primitives via a geometric
retention score; surface alignment regularizes each primitive into a plate aligned with
the local surface; and outward normal assignment assigns each primitive an outward-pointing normal.

\subsubsection{Geometry-Driven Densification}
\label{sec:gaussian:density}

Standard 3DGS optimizes $T$ anisotropic primitives by photometric loss~\cite{kerbl20233d},
using gradient-driven adaptive density control to clone or split primitives
with large view-space positional gradients and prune low-opacity ones. However, under appearance-based supervision,
these gradients may be dominated by texture rather than
geometry. Primitives could be over-allocated to texturally rich but
geometrically simple regions, while redundant floaters or overlapping splats
may survive as long as their opacity remains above the pruning threshold.

We instead drive densification with mesh-rendered normal supervision. For each
view, we encode the surface normals as RGB channels to form the target image
$C^{\mathrm{gt}}$, with $\hat{C}$ the corresponding rendering from the Gaussian
primitives. This redirects the optimization from reconstructing appearance to
reconstructing surface geometry: since normals are encoded as color, curved
regions yield sharp color gradients, whereas planar regions stay
nearly uniform. Since densification is triggered by accumulated image-space
gradients, this supervision produces more, smaller primitives in geometrically
complex regions and fewer, larger ones in simple regions.

To further strengthen densification in geometrically complex regions, we weight the standard 3DGS $\mathcal{L}_1$ photometric term by a curvature-aware map $W$ from the
mesh-rendered depth map $D^{\mathrm{gt}}$:
{
\compacteqspacing
\begin{equation}
W(u,v) = 1 + \mathrm{Norm}\big(|\partial_u D^{\mathrm{gt}}(u,v)|
+ |\partial_v D^{\mathrm{gt}}(u,v)|\big),
\end{equation}
}
where $(u,v)$ indexes pixels, $\partial_u,\partial_v$ are image-axis finite
differences, and $\mathrm{Norm}(\cdot)$ rescales the depth gradient to $[0,1]$ over valid pixels.
The unit offset shifts the range of $W$ to $[1,2]$, ensuring that all valid pixels contribute to the weighted $\mathcal{L}_1$ photometric term.
The resulting map $W$ up-weights reconstruction errors at depth discontinuities and
sharp transitions.
With $(\lambda_{\mathrm{C}},\lambda_{\mathrm{S}})=(0.8,0.2)$, the photometric loss is:
{
\compacteqspacing
\begin{equation}
\mathcal{L}_{\mathrm{photo}} = \lambda_{\mathrm{C}}
\big\| W \odot |\hat{C}-C^{\mathrm{gt}}| \big\|_1
+ \lambda_{\mathrm{S}} \mathcal{L}_{\mathrm{D\text{-}SSIM}}(\hat{C}, C^{\mathrm{gt}}),
\end{equation}} 
where
$\mathcal{L}_{\mathrm{D\text{-}SSIM}}(\hat{C},C^{\mathrm{gt}})
=
1-\mathrm{SSIM}(\hat{C},C^{\mathrm{gt}})$
denotes the structural dissimilarity loss based on the Structural
Similarity Index Measure (SSIM)~\cite{wang2004image}.

To remove redundant primitives, we add a soft opacity-sparsity term and
optimize the objective at this stage:
{
\compacteqspacing
\begin{equation}
\mathcal{L}_{\mathrm{geo}}
= \mathcal{L}_{\mathrm{photo}} + \lambda_{\mathrm{opa}}\mathcal{L}_{\mathrm{opa}},
\qquad
\mathcal{L}_{\mathrm{opa}} = \frac{1}{T}\sum_{i=1}^{T} o_i,
\end{equation}
}
where $o_i$ is the opacity of the $i$-th primitive and
$\lambda_{\mathrm{opa}}=0.02$. Under this term, redundant floaters and
overlapping primitives are driven toward transparency and removed by the
standard opacity-pruning mechanism, whereas primitives essential for normal-view reconstruction retain sufficient opacity through $\mathcal{L}_{\mathrm{photo}}$. $\mathcal{L}_{\mathrm{opa}}$ is enabled only after a short warm-up period. Consequently,
sharp and highly curved regions receive fine-scale Gaussian
coverage, whereas smooth, approximately planar regions
retain fewer non-redundant primitives that can cover broader
surface patches.

\subsubsection{Fixed-Budget Selection}
\label{sec:gaussian:realloc}
To obtain a fixed-budget primitive set for the downstream model, we retain the top-$G$ primitives according to $\mathrm{score}(g_j)=o_j A_j(1+\beta H_j),
$ where $A_j$ is the maximum projected area, and $H_j\in[0,1]$ is the normalized discrete mean curvature of the nearest mesh vertex to $\boldsymbol{\mu}_j$. 
The term $o_jA_j$ favors primitives with substantial surface coverage, while $H_j$ favors primitives in high-curvature regions.
We empirically set $\beta=0.2$.
\subsubsection{Surface Alignment Refinement}
\label{sec:gaussian:plate}
Inspired by~\cite{guédon2023sugarsurfacealignedgaussiansplatting,huang20242d}, we refine each retained primitive into a thin plate that adheres to the object surface and aligns with its local geometry.
With the budget fixed at $G$ primitives, densification and culling are disabled, so optimization is restricted entirely to the geometric parameters $\{\boldsymbol{\mu}_j, \mathbf{R}_j, \boldsymbol{\sigma}_j\}_{j=1}^{G}$ under the objective:
\begin{equation}
\label{eq:stage2}
\mathcal{L}_{\mathrm{refine}} = \mathcal{L}_{\mathrm{photo}} + \lambda_s \mathcal{L}_{\mathrm{surf}} + \lambda_n \mathcal{L}_{\mathrm{norm}} + \lambda_p \mathcal{L}_{\mathrm{plate}},
\end{equation}
where $\mathcal{L}_{\mathrm{photo}}$ is retained to further drive existing primitives to translate and expand across unrepresented surface regions. The remaining three geometric regularizers act on each primitive against its closest sampled point $\mathbf{x}_j$ on the source mesh and the mesh surface normal $\mathbf{n}^{\mathrm{m}}_j$ ($\mathrm{m}$ for \emph{mesh}) at $\mathbf{x}_j$.

\textit{Surface attraction} pulls each primitive center $\boldsymbol{\mu}_j$ to its closest mesh point $\mathbf{x}_j$:
{
\compacteqspacing
\begin{equation}
\mathcal{L}_{\mathrm{surf}} =  \sum_j \| \boldsymbol{\mu}_j - \mathbf{x}_j \|^2.
\end{equation}
}

\textit{Normal alignment} forces the shortest principal axis direction $\mathbf{n}^{\mathrm{s}}_j$ ($\mathrm{s}$ for \emph{shortest principal axis}) of the primitive to remain parallel to $\mathbf{n}^{\mathrm{m}}_j$:
{
\compacteqspacing
\begin{equation}
\mathcal{L}_{\mathrm{norm}} =  \sum_j \big(1 - |\mathbf{n}^{\mathrm{s}}_j \cdot \mathbf{n}^{\mathrm{m}}_j|\big).
\end{equation}
}

\textit{Plate regularization} acts on the sorted primitive scales $s^{(1)}_j \leq s^{(2)}_j \leq s^{(3)}_j$ of
$\boldsymbol{\sigma}_j$ through one-sided hinge penalties, $[x]_+ = \max(0, x)$:
{
\compacteqspacing
\begin{equation}
\mathcal{L}_{\mathrm{plate}}\!=\!\sum_j\!\Big(
[s_j^{(1)}\!-\!\tau_h]_+
+[\tau_\ell\!-\!s_j^{(1)}]_+
+\sum_{k=2}^{3}[\tau_f\!-\!s_j^{(k)}]_+
\Big),
\end{equation}
}
where the first two terms confine the primitive thickness $s^{(1)}_j$ to a narrow band $[\tau_\ell, \tau_h]$. The final term floors the in-plane axes at $\tau_f$, encouraging the plates to stretch tangentially along the surface to broaden geometric coverage.

\subsubsection{Outward Normal Assignment}
The normal-alignment loss $\mathcal{L}_{\mathrm{norm}}$ aligns the shortest
principal axis $\mathbf{n}^{\mathrm{s}}_j$ with the local surface normal but
does not determine its sign. After refinement, we project each primitive center
$\boldsymbol{\mu}_j$ onto the source mesh and obtain the outward normal
$\mathbf{n}^m_j$ of the closest surface element. We then set
$\mathbf{n}_j =
\mathrm{sgn}(\mathbf{n}^{\mathrm{s}}_j \cdot \mathbf{n}^m_j)
\mathbf{n}^{\mathrm{s}}_j$, to ensure that all primitive normals point outward.

\subsection{Pretraining Morpho-Kinematic Robot Descriptors}
\label{sec:pretraining}
\begin{figure}[t]
    \centering
    \includegraphics[
        width=0.95\linewidth,
    ]{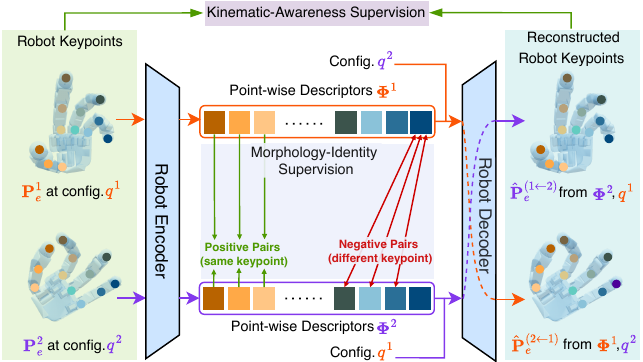}
    \vspace{-10pt}
    \caption{
    Pretraining of morpho-kinematic robot descriptors. For two configurations
    $\mathbf{q}^{1}$ and $\mathbf{q}^{2}$ of the same embodiment, a robot
    encoder maps the corresponding keypoint sets $\mathbf{P}_e^{1}$ and
    $\mathbf{P}_e^{2}$ to point-wise descriptors $\Phi^{1}$ and $\Phi^{2}$.
    The Morphology-Identity Supervision aligns descriptors of the same physical
    keypoints across configurations while separating those of different keypoints.
    Kinematic-Awareness Supervision uses a robot decoder to reconstruct  
    $\hat{\mathbf{P}}_e^{(2\leftarrow 1)}$ from $\Phi^{1}$ and $\mathbf{q}^{2}$,
    and $\hat{\mathbf{P}}_e^{(1\leftarrow 2)}$ from $\Phi^{2}$ and
    $\mathbf{q}^{1}$, minimizing errors to the original keypoint sets
    $\mathbf{P}_e^{2}$ and $\mathbf{P}_e^{1}$, respectively.
    }
    \vspace{-10pt}
    \label{fig:pretrain}
\end{figure}
We pretrain a robot encoder to produce morpho-kinematic descriptors for hand keypoints. For dexterous grasping, each descriptor should encode configuration-invariant morphological identity while remaining predictive of keypoint motion under changes in hand configuration. To this end, we combine a morphology-identity contrastive objective with a kinematic awareness objective, as illustrated in Fig.~\ref{fig:pretrain}.

\noindent\textbf{Pretraining Setup.} For each pretraining instance, we sample an embodiment $e$ and two hand configurations
$\mathbf{q}^{1}$ and $\mathbf{q}^{2}$. The corresponding keypoint sets
$\mathbf{P}_e^{1}$ and $\mathbf{P}_e^{2}$ are obtained by forward kinematics. A  DGCNN-based robot encoder~\cite{wang2019dynamic} $E_{\mathrm{robot}}$ maps each keypoint set to point-wise descriptors:
$\Phi^{k}
=
E_{\mathrm{robot}}(\mathbf{P}_e^{k})
=
\{\boldsymbol{\phi}^{k}_i\in\mathbb{R}^{512}\}_{i=1}^{N},
\quad k\in\{1,2\}.$

\noindent\textbf{Morphology-Identity Supervision.} We follow the geometry-aware contrastive formulation of~\cite{11127754DRO}. To preserve morphological identity, descriptors of the same physical keypoint should remain consistent across configurations, while descriptors of different keypoints should remain distinguishable.
With descriptor similarity
$S_{ij}=
\frac{\langle\boldsymbol{\phi}^{1}_i,\boldsymbol{\phi}^{2}_j\rangle}
{\|\boldsymbol{\phi}^{1}_i\|_2\|\boldsymbol{\phi}^{2}_j\|_2}$
and distance
$d_{ij}=\|\mathbf{p}^{1}_i-\mathbf{p}^{1}_j\|_2$,
the contrastive logit is:

{\compacteqspacing
\begin{equation}
z_{ij}
=
\log\!\left(
\frac{\tanh d_{ij}}{\max_k \tanh d_{ik}}
+
\delta_{ij}
\right)
+
\frac{S_{ij}}{\tau},
\end{equation}} where $\delta_{ij}$ is the Kronecker delta and $\tau=0.1$ is the temperature. Further, the morphology-identity loss is:
{\compacteqspacing
\begin{equation}
\mathcal{L}_{\mathrm{morph}}
=
-\sum_i
\log
\frac{\exp(z_{ii})}
{\sum_j\exp(z_{ij})}.
\end{equation}
}
\noindent\textbf{Kinematic-Awareness Supervision.} To make descriptors predictive of keypoint motion under changes in hand configuration, we reuse the same configuration pair and impose a cross-conditioned prediction task. Given descriptors extracted from a source configuration and the target hand configuration, robot decoder $D_{\mathrm{fk}}$ predicts the target keypoint locations:
$
\hat{\mathbf{P}}_e^{(2\!\leftarrow\!1)}\!=\!D_{\mathrm{fk}}(\Phi^{1},\mathbf{q}^{2}),\quad
\hat{\mathbf{P}}_e^{(1\!\leftarrow\!2)}\!=\!D_{\mathrm{fk}}(\Phi^{2},\mathbf{q}^{1})$.
Since grasp stability is dominated by contact-relevant regions, we further weight the regression by an offline, embodiment-level contact prior. For each embodiment $e$, near-contact frequency statistics are accumulated over hand-mesh vertices from the training grasps and propagated to keypoints by nearest-neighbor association in the local link frame. Letting $c^{e}_i\in[0,1]$ denote this prior for the $i$-th keypoint of embodiment $e$, per-keypoint weights are:
{\compacteqspacing
\begin{equation}
\label{contact_weights}
w^{e}_i
=\frac{c^{e}_i+\alpha}{\tfrac{1}{N}\sum_{j=1}^{N}(c^{e}_j+\alpha)},
\end{equation}}
where $\alpha=0.1$ floors near-zero weights. The contact-aware kinematic-awareness loss is then:
{
\compacteqspacing
\begin{equation}
\mathcal{L}_{\mathrm{kin}}
=
\sum_{i=1}^{N} w^{e}_i
\big(\|\hat{\mathbf{p}}^{(1\leftarrow 2)}_i-\mathbf{p}^{(1)}_i\|_2^2
    +\|\hat{\mathbf{p}}^{(2\leftarrow 1)}_i-\mathbf{p}^{(2)}_i\|_2^2\big).
\end{equation}
}
\noindent\textbf{Overall Objective.} The two pretraining terms jointly learn morpho-kinematic descriptors, giving the final objective:
{
\compacteqspacing
\begin{equation}
\mathcal{L}_{\mathrm{pre}}
=
\mathcal{L}_{\mathrm{morph}}
+
\mathcal{L}_{\mathrm{kin}}.
\end{equation}
}
\subsection{Interaction Fields Generation}
\label{sec:prediction}
\subsubsection{Modeling Interaction with Mahalanobis Fields}
We define the interaction fields between the robot and the object as a pairwise Mahalanobis distance~\cite{mahalanobis2018generalized} matrix $\mathbf{M}\in \mathbb{R}^{N \times G}$ measured between robot keypoints and object primitives. For a keypoint $\mathbf{p}_i$ and primitive $g_j = (\boldsymbol{\mu}_j, \mathbf{R}_j, \boldsymbol{\sigma}_j)$, the field is
{
\compacteqspacing
\begin{equation}
\label{eq:mahalanobis}
M_{ij} = \sqrt{\big(\mathbf{R}_j^\top(\mathbf{p}_i \!-\! \boldsymbol{\mu}_j)\big)^\top \mathrm{diag}(\boldsymbol{\sigma}_j^{-2}) \big(\mathbf{R}_j^\top(\mathbf{p}_i \!-\! \boldsymbol{\mu}_j)\big)}.
\end{equation}
}

Since our primitives are surface-aligned plates, with broad faces tangent to the surface and short axes along the surface normal, the resulting fields are more sensitive to offsets along the surface-normal direction than along tangential directions. The field's sensitivity to tangential offsets also varies with the primitive's tangential scale, with smaller scale yielding greater sensitivity.

\subsubsection{Network Architecture}
Initialized robot keypoints $\mathbf{P}^{0}_e$ are encoded by the frozen
pretrained robot encoder to obtain Keypoint-wise Robot Descriptors
$\mathbf{F}^R\in\mathbb{R}^{N\times d}$. And the object branch embeds
each Gaussian primitive into a feature vector and builds a KNN graph over
primitive centers. Stacked local attention blocks refine these features using
neighborhood edge features that encode relative position and scale between
neighboring primitives. A global self-attention layer then mixes information across all primitives, injecting object-level context into each token, producing Primitive-wise Object Features $\mathbf{F}^O\in\mathbb{R}^{G\times d}$.

Given $\mathbf{F}^R$ and $\mathbf{F}^O$, the Interaction Fields Generator generates $\hat{\mathbf{M}}$. Following~\cite{11127754DRO}, it first contextualizes the two streams with cross-attention, yielding $\tilde{\mathbf{F}}^R$ and $\tilde{\mathbf{F}}^O$. It then samples a latent code $\mathbf{z}\in\mathbb{R}^{d_z}$ from a CVAE~\cite{sohn2015learning}, drawing from the posterior in training and from $\mathcal{N}(\mathbf{0},\mathbf{I})$ at inference. Finally, for each pair $(i,j)$, it concatenates $\tilde{\mathbf{F}}^R_i$ and $\tilde{\mathbf{F}}^O_j$ with $\mathbf{z}$, and passes the result through a non-negative pairwise MLP to predict $\hat{M}_{ij}$, giving $\hat{\mathbf{M}}\in\mathbb{R}^{N\times G}$ over all $N\times G$ pairs.

\subsubsection{Training Objective}
We supervise the network against the ground truth
$\mathbf{M}^*$, computed by evaluating Eq.~\eqref{eq:mahalanobis} between the grasp-state hand keypoints and the corresponding object Gaussian primitives. Since the resulting targets are heavy-tailed: contact pairs cluster near zero while distant pairs span much larger values, we apply a $\log(1+x)$ compression
to both prediction and target. We also reuse the contact-aware keypoint
weights $w^{e}_i$ from Sec.~\ref{sec:pretraining} to concentrate
supervision on grasp-relevant keypoints:
{
\compacteqspacing
\begin{equation}
\mathcal{L}_{\mathrm{reg}}
= \frac{1}{NG}\sum_{i,j} w^{e}_i\,
  \big|\log(1+\hat{M}_{ij}) - \log(1+M^*_{ij})\big|.
\end{equation}
}
A KL term $\mathcal{L}_{\mathrm{kl}}$ regularizes the CVAE posterior. Using
$\lambda_{\mathrm{reg}}=10$ and $\lambda_{\mathrm{kl}}=0.01$, we optimize:
{
\compacteqspacing
\begin{equation}
\mathcal{L}
=
\lambda_{\mathrm{reg}}\mathcal{L}_{\mathrm{reg}}
+
\lambda_{\mathrm{kl}}\mathcal{L}_{\mathrm{kl}}.
\end{equation}}
\subsection{Grasp Realization via Optimization}
\label{sec:recovery}
We initialize optimization from an object-conditioned pre-grasp
configuration $\mathbf{q}^{0}$. Specifically, we sample an object primitive
$g_j$, place the wrist anchor at an offset from $\boldsymbol{\mu}_j$ along the
outward normal $\mathbf{n}_j$, and orient the hand approach direction toward
the object with a random roll around $\mathbf{n}_j$. The finger joints are
initialized to an open configuration. Based on the predicted target Mahalanobis field $\hat{\mathbf{M}}$, we realize the final grasp $\mathbf{q}^*$ by optimizing the hand configuration $\mathbf{q}$:
{\compacteqspacing
\begin{equation}
\label{eq:optimization}
\min_{\mathbf{q}}\ \lambda_{\mathrm{guide}}\mathcal{L}_{\mathrm{guide}}+\lambda_{\mathrm{pen}}\mathcal{L}_{\mathrm{pen}}+\lambda_{\mathrm{self}}\mathcal{L}_{\mathrm{self}},
\end{equation}}
and the revolute joints  $\mathbf{q}_r$ are constrained to joint limit $[\underline{\mathbf{q}}_r,\overline{\mathbf{q}}_r]$.
\emph{Requiring no embodiment-specific retuning}, we use one weight setting
$(\lambda_{\mathrm{guide}},\lambda_{\mathrm{pen}},\lambda_{\mathrm{self}})=(50,60,10)$ for all hands.
The remaining three terms are defined as follows.

\noindent\textbf{Mahalanobis Fields Guidance.}
$\mathcal{L}_{\mathrm{guide}}$ guides the current Mahalanobis fields toward the prediction. Since the predicted fields are learned from stable grasp examples, this guidance serves as a data-driven geometric prior for interaction patterns associated with stable grasps.
The geometry-oriented 3D Gaussian representation
also allows the guidance to account for local
geometric complexity. 
Smaller primitives in geometrically complex regions can impose tighter tangential constraints, consistent with the smaller surface patches they represent, 
while broader primitives in simpler
regions allow greater tangential adjustment. Moreover, denser primitive allocation in geometrically complex regions provides more local guidance terms that can jointly constrain keypoint motion. The contact-aware keypoint weights $w^{e}_i$ from Sec.~\ref{sec:pretraining} are used to focus the guidance on contact-relevant keypoints:
{
\compacteqspacing
\begin{equation}
\mathcal{L}_{\mathrm{guide}} \!=\! \frac{1}{NG} \!\sum_{i,j}\! w^{e}_i \!\cdot\! \psi_g\!\Big(\!\log\!\big(1 \!+\! M_{ij}\big),\! \log\!\big(1 \!+\! \hat{M}_{ij}\big)\!\Big)\!,
\end{equation}
}
where $\log(1+x)$ compression is used to both fields for stability. $\psi_g(a,b)=[a-b]_+$ penalizes the current field only when it exceeds the prediction. This leaves the optimizer free to pull in tighter than predicted when the geometry permits, with over-closure handled by the penetration term $\mathcal{L}_{\mathrm{pen}}$.

\noindent\textbf{Penetration Energy.}
$\mathcal{L}_{\mathrm{pen}}$ penalizes hand-object penetration using object Gaussian primitives. For each hand keypoint $\mathbf{p}_i$, we select its nearest primitive under the Mahalanobis metric,
$j^*(i)=\arg\min_j M_{ij}$, and compute its signed normal offset
$s_i=(\mathbf{p}_i-\boldsymbol{\mu}_{j^*(i)})^\top \mathbf{n}_{j^*(i)}$. Such penetration estimation also benefits from the finer
local surface approximations provided by our
geometry-oriented 3D Gaussian representation.
{
\compacteqspacing
\begin{equation}
\mathcal{L}_{\mathrm{pen}}
=
\operatorname{TopKSum}_{i,K}\!\left[
\psi_p(s_i)
\right],
\end{equation}
}
where $\psi_p(s)=(m-s)_+$, and $\operatorname{TopKSum}_{i,K}$ sums the largest $K$ values over $i$. We set $m=-0.001$ and $K=20$.

\noindent\textbf{Self-Collision Energy.}
$\mathcal{L}_{\mathrm{self}}$ penalizes close keypoint pairs on non-adjacent links:
{
\compacteqspacing
\begin{equation}
\hspace*{-0.3em}
\mathcal{L}_{\mathrm{self}}
=
\operatorname{TopKSum}_{i,K}\!\left[
\max_{{j:\,d_{k}(\ell_i,\ell_j)>1}}
\psi_s(\|\mathbf{p}_i-\mathbf{p}_j\|_2)
\right],
\end{equation}
}
where $\psi_s(r)=(\epsilon_s-r)_+$, $d_{k}$ denotes graph distance in the hand kinematic tree, and $\operatorname{TopKSum}_{i,K}$ sums the largest $K$ values over $i$. We set $\epsilon_s=0.01$ and $K=20$.

\section{Experiments}
Experiments are designed to answer the following questions: \textbf{Q1}) How does MANGO-Grasp compare with existing cross-embodiment grasp synthesis baselines? \textbf{Q2}) Does it generalize zero-shot to unseen robotic hands? 
\textbf{Q3}) Does morpho-kinematic pretraining improve over pretraining
with only morphology-identity supervision (as defined in Sec.~\ref{sec:pretraining})? \textbf{Q4}) How does the proposed Mahalanobis Field compare with Euclidean fields for contact representation? \textbf{Q5}) How much does the 3D Gaussian representation improve over a point-cloud alternative, and \textbf{Q6}) does it outperform mesh-fitted primitives? 
\textbf{Q7}) What is the effect of the contact prior during model training and grasp realization?
\textbf{Q8}) How does MANGO-Grasp perform in the real world?

\subsection{Experimental Setting}
\label{datasets}
\noindent\textbf{Implementation Details.}
We implement the model in PyTorch. With 29.5M parameters, it is trained for 150 epochs on an NVIDIA RTX 5090 GPU, requiring approximately 7.4 hours. We use Adam optimizer with an initial learning rate of $10^{-5}$, a learning rate scheduler and a batch size of 24. 
The average inference time
of the full pipeline is $0.124~\mathrm{s}$ per grasp,
including $0.013~\mathrm{s}$ for interaction-field prediction
and $0.111~\mathrm{s}$ for iterative grasp optimization.
And the offline Gaussian construction takes
$47.8~\mathrm{s}$ per object on average.

\noindent\textbf{Datasets.}
Two datasets are used for benchmark in this work. 1) We adopt the filtered CMAP dataset~\cite{li2023gendexgrasp} with DRO~\cite{11127754DRO}'s train/test split, which is also consistent with TRO~\cite{fei2025trograspefficientgraph}. The training split contains 14{,}011 grasps over 48 objects and three morphologically distinct hands (ShadowHand, Allegro, and Barrett), and the test split contains 10 novel objects. 2) To assess object-level generalization, we further
evaluate on a randomly selected 40-object subset of MultiGripperGrasp~\cite{casas2024multigrippergrasp}
that spans broader geometries and scales.

\noindent\textbf{Simulation Setup.}
We evaluate grasp stability in Isaac Gym~\cite{liang2018gpu} using the PhysX engine. Following the simulation protocol in prior works~\cite{li2023gendexgrasp,11127754DRO}, each hand is initialized at the predicted pose with the target object present. External forces of magnitude $0.5m_og$ where $m_o$ is the object mass, are applied sequentially to the object along six orthogonal directions. A grasp is deemed successful if hand--object contact is preserved after the full perturbation sequence.

\noindent\textbf{Baselines.}
We include four baselines. \textit{DexGraspNet}~\cite{wang2022dexgraspnet} optimizes differentiable force closure~\cite{liu2021synthesizing} and kinematic feasibility; as its implementation supports only ShadowHand, we extend it to Allegro and Barrett with careful hyperparameter tuning and report the best-performing results among multiple settings.
\textit{DRO}~\cite{11127754DRO} predicts distances between robot keypoints and object point clouds to guide grasp optimization. \textit{TRO}~\cite{fei2025trograspefficientgraph} is a leading open-source method that diffuses SE(3) transformations over hand-link and object-patch graphs to guide joint optimization.
\textit{CEDex}~\cite{wu2025cedexcrossembodimentdexterousgrasp} adapts human-like contact representations to robot hands and realizes grasps through optimization; as SharpaWave is not included in its implementation, we implement it following the paper and the provided implementations for other hands.
For fairness, we retrain \emph{DRO}, \emph{TRO}, and \emph{CEDex} and report the reproduced results.

\subsection{Main Results}
For each method, we generate 100 candidate grasps for each test object and report the mean simulation success rate with its two-sided 95\% confidence interval across three independent runs, computed using Student's $t$ distribution.

\begin{table*}[!t]
\centering

\begingroup

\caption{Simulation success rates (\%) for baseline comparison and ablation studies across seen hands and unseen SharpaWave.}
\vspace{-10pt}
\label{tab:main_result}

\setlength{\tabcolsep}{3.2pt}
\renewcommand{\arraystretch}{1}
\small

\resizebox{\textwidth}{!}{%
\begin{tabular}{@{}l | cccc | c | cccc | c@{}}
\hline
\multirow{3}{*}{\textbf{Method / Variant}}
& \multicolumn{5}{c|}{\textbf{CMAP Test Split}}
& \multicolumn{5}{c}{\textbf{MultiGripperGrasp Test Split}} \\
\cline{2-11}
& \multicolumn{4}{c|}{\textit{Seen Hands}} & \textit{Unseen Hand}
& \multicolumn{4}{c|}{\textit{Seen Hands}} & \textit{Unseen Hand} \\
\cline{2-5} \cline{6-6} \cline{7-10} \cline{11-11}
& ShadowHand & Allegro & Barrett & \textbf{Avg.} & SharpaWave
& ShadowHand & Allegro & Barrett & \textbf{Avg.} & SharpaWave \\
\hline

DexGraspNet~\cite{wang2022dexgraspnet}
& $69.23_{\pm 1.17}$
& $62.77_{\pm 1.25}$
& $70.63_{\pm 0.94}$
& $67.54_{\pm 0.19}$
& --$^{*}$
& $62.48_{\pm 2.17}$
& $59.38_{\pm 2.53}$
& $65.72_{\pm 1.87}$
& $62.53_{\pm 1.09}$
& --$^{*}$ \\

DRO~\cite{11127754DRO}
& $82.70_{\pm 1.08}$
& $92.10_{\pm 2.12}$
& $88.53_{\pm 2.49}$
& $87.78_{\pm 0.91}$
& $70.70_{\pm 2.77}$
& $70.45_{\pm 2.42}$
& $76.87_{\pm 1.72}$
& $78.88_{\pm 1.03}$
& $75.40_{\pm 1.71}$
& $64.90_{\pm 2.52}$ \\

TRO~\cite{fei2025trograspefficientgraph}
& $95.23_{\pm 0.94}$
& $94.50_{\pm 1.38}$
& $92.90_{\pm 2.17}$
& $94.21_{\pm 0.56}$
& $39.33_{\pm 2.88}$
& $75.65_{\pm 1.76}$
& $80.81_{\pm 3.09}$
& $87.22_{\pm 1.06}$
& $81.23_{\pm 1.90}$
& $24.98_{\pm 2.30}$ \\

CEDex~\cite{wu2025cedexcrossembodimentdexterousgrasp}
& $84.43_{\pm 2.07}$
& $92.77_{\pm 1.17}$
& $91.30_{\pm 0.66}$
& $89.50_{\pm 0.74}$
& $62.60_{\pm 4.13}$
& $75.85_{\pm 0.35}$
& $80.05_{\pm 1.08}$
& $86.97_{\pm 1.58}$
& $80.96_{\pm 0.43}$
& $64.45_{\pm 0.55}$ \\

\hline

\textbf{MANGO-Grasp (full)}
& $\mathbf{96.50}_{\pm 1.38}$
& $\mathbf{98.47}_{\pm 0.14}$
& $\mathbf{97.80}_{\pm 0.25}$
& $\mathbf{97.59}_{\pm 0.37}$
& $\mathbf{84.17}_{\pm 2.14}$
& $\mathbf{85.97}_{\pm 1.55}$
& $\mathbf{93.12}_{\pm 1.02}$
& $\mathbf{89.33}_{\pm 0.85}$
& $\mathbf{89.47}_{\pm 0.39}$
& $\mathbf{81.47}_{\pm 2.36}$ \\

w/o Kinematic Awareness
& $87.03_{\pm 1.27}$
& $91.90_{\pm 1.14}$
& $88.50_{\pm 0.43}$
& $89.14_{\pm 0.60}$
& $77.53_{\pm 2.79}$
& $80.88_{\pm 1.48}$
& $89.28_{\pm 0.66}$
& $85.69_{\pm 2.41}$
& $85.28_{\pm 0.47}$
& $74.50_{\pm 2.27}$ \\

Euclidean Fields
& $79.57_{\pm 1.15}$
& $84.97_{\pm 1.99}$
& $82.87_{\pm 0.52}$
& $82.47_{\pm 0.30}$
& $65.53_{\pm 1.65}$
& $72.49_{\pm 1.20}$
& $80.02_{\pm 1.96}$
& $76.35_{\pm 1.47}$
& $76.29_{\pm 1.17}$
& $61.50_{\pm 1.77}$ \\

Point-Cloud Representation
& $69.27_{\pm 2.01}$
& $75.23_{\pm 1.43}$
& $70.93_{\pm 0.38}$
& $71.81_{\pm 0.75}$
& $54.10_{\pm 2.37}$
& $64.43_{\pm 1.62}$
& $68.63_{\pm 1.32}$
& $66.41_{\pm 1.53}$
& $66.49_{\pm 1.35}$
& $51.63_{\pm 1.90}$ \\

Mesh-Fitted Primitives
& $81.80_{\pm 2.03}$
& $87.23_{\pm 1.86}$
& $83.43_{\pm 1.00}$
& $84.16_{\pm 0.46}$
& $74.07_{\pm 2.01}$
& $76.99_{\pm 2.75}$
& $85.77_{\pm 1.02}$
& $80.62_{\pm 1.33}$
& $81.13_{\pm 1.42}$
& $67.90_{\pm 2.44}$ \\

w/o Contact Prior
& $91.90_{\pm 0.66}$
& $94.23_{\pm 0.38}$
& $93.03_{\pm 1.90}$
& $93.06_{\pm 0.75}$
& $80.77_{\pm 1.46}$
& $81.32_{\pm 1.69}$
& $91.44_{\pm 0.95}$
& $86.80_{\pm 1.89}$
& $86.52_{\pm 0.84}$
& $79.23_{\pm 3.54}$ \\

Contact Prior (Realization Only)
& $93.30_{\pm 1.88}$
& $95.87_{\pm 1.41}$
& $93.77_{\pm 1.37}$
& $94.31_{\pm 0.77}$
& --$^{\dagger}$
& $83.30_{\pm 1.71}$
& $92.55_{\pm 0.19}$
& $87.59_{\pm 1.17}$
& $87.81_{\pm 0.88}$
& --$^{\dagger}$ \\

Contact Prior (Training Only)
& $95.03_{\pm 0.29}$
& $96.47_{\pm 0.29}$
& $96.10_{\pm 0.50}$
& $95.87_{\pm 0.29}$
& $84.17_{\pm 2.14}$$^{\S}$
& $82.97_{\pm 0.69}$
& $92.20_{\pm 0.51}$
& $88.96_{\pm 1.52}$
& $88.04_{\pm 0.58}$
& $81.47_{\pm 2.36}$$^{\S}$ \\

\hline
\end{tabular}
}

\vspace{-4pt}
\begin{flushleft}
\footnotesize
$^{*}$DexGraspNet is optimization-based and does not involve learned embodiment transfer, so it is excluded from the zero-shot unseen-hand comparison.

$^{\dagger}$SharpaWave has no contact prior under the unseen-hand setting, the experimental setting that requires a contact prior during realization is not applicable.

$^{\S}$For SharpaWave, the training-only setting is identical to the full model setup because both use uniform weights during grasp realization; therefore, we reuse the result from the full-model setup.
\end{flushleft}

\endgroup
\vspace{-20pt}
\end{table*}

Table~\ref{tab:main_result} shows that MANGO-Grasp achieves the highest
seen-hand average on both unseen-object benchmarks, reaching 97.59\% on CMAP
and 89.47\% on MultiGripperGrasp. Among the baselines, TRO achieves the best
seen-hand average on both benchmarks. MANGO-Grasp further improves this average
by 3.38 and 8.24 percentage points (pp) on CMAP and MultiGripperGrasp, respectively.
The gains are consistent across ShadowHand, Allegro, and Barrett, indicating
robust grasp synthesis across seen embodiments.

MultiGripperGrasp is more challenging due to broader object geometry and scale
variation. From CMAP to MultiGripperGrasp, DRO, TRO, and CEDex drop by 12.38, 12.98, and 8.54 pp, respectively, while MANGO-Grasp drops by 8.12 pp. This smaller degradation suggests stronger object-level generalization.
DexGraspNet yields more modest results than the learning-based methods despite extensive
per-hand hyperparameter tuning, reflecting the sensitivity of optimization-based
methods to embodiment-specific tuning.
\subsection{Zero-Shot Generalization to Unseen Hand}
\label{sec:unseen_hand}
To address \textbf{Q2}, we assess zero-shot embodiment generalization by
testing the learning-based methods on the unseen SharpaWave hand with complex morphology and kinematics (5 fingers with 22 joints).

Since no contact prior is available for the unseen SharpaWave, we assign equal importance to all
keypoints, yielding $w_i^e=1$
under the unit-mean normalization of Eq.~(\ref{contact_weights}).
As shown in Table~\ref{tab:main_result}, MANGO-Grasp
outperforms the strongest zero-shot baseline DRO by 13.47 pp on CMAP and 16.57 pp on MultiGripperGrasp,
respectively. CEDex shows reduced performance on SharpaWave, reaching 62.60\% and 64.45\%. TRO performs strongly on seen hands but degrades on SharpaWave,
consistent with its reported limitation that zero-shot generalization is limited to hands with similar morphology. This suggests that MANGO-Grasp learns a more transferable interaction representation.
\subsection{Ablation Studies}
\label{sec:ablation}
To answer \textbf{Q3}--\textbf{Q7}, we ablate the core components in our pipeline. And each variant's results are reported in  Table~\ref{tab:main_result}.

\noindent\textbf{Effect of the Kinematic-Awareness Supervision (Q3).}
We remove the kinematic-awareness supervision from robot pretraining, leaving only the morphology-identity supervision. This reduces the seen-hand average by 8.45 and 4.19 pp on CMAP and MultiGripperGrasp, and decreases zero-shot SharpaWave
performance by 6.64 and 6.97 pp. 
These results highlight the effect of the kinematic-awareness supervision.

\noindent\textbf{Mahalanobis Fields vs.\ Euclidean Fields (Q4).}
We replace the Mahalanobis prediction target with Euclidean Fields between robot keypoints and Gaussian primitive centers, and realize grasps from the isotropic Euclidean Fields. On seen hands, the average drops by 15.12 pp on CMAP and 13.18 pp on MultiGripperGrasp; unseen SharpaWave drops by 18.64 and 19.97 pp. This shows that Mahalanobis Fields provide a more effective surface-aware signal than Euclidean Fields.

\noindent\textbf{3D Gaussians vs.\ Point Clouds (Q5).}
We replace the 3D Gaussians with point clouds. Unlike Gaussians, point clouds offer no explicit surface geometry encoding and lack adaptive density. They also carry no covariance, so the interaction fields have to reduce to Euclidean fields. This variant yields seen-hand average drops of 25.78 and 22.98 pp on CMAP and MultiGripperGrasp, and zero-shot SharpaWave drops of 30.07 and 29.84 pp, respectively. The result reflects two roles of the 3D Gaussian: 
it provides richer local surface information than the point-cloud alternative,
and its local surface encoding enables anisotropic Mahalanobis fields.

\noindent\textbf{3D Gaussians vs.\ Mesh-Fitted Primitives (Q6).}
We replace the 3D Gaussians with anisotropic primitives directly built from the mesh using curvature-adaptive sampling and quadric fitting. This variant yields 13.43 and 8.34 pp drops on seen hands for CMAP and MultiGripperGrasp, respectively, and 10.10 and 13.57 pp drops on zero-shot SharpaWave, showing that our 3DGS-based construction provides gains beyond oriented anisotropic surface primitives alone.

\noindent\textbf{Effect of the Contact Prior (Q7).}
We ablate contact prior during model training
and grasp realization under three variants: 1) \textit{w/o Contact Prior}: we set $w_i^e{=}1$ during both training and grasp realization,
which reduces the seen-hand average by 4.53 and 2.95 pp on CMAP and
MultiGripperGrasp, respectively. For unseen SharpaWave, this variant
also reduces performance by 3.40 and 2.24 pp, respectively;
2) \textit{Contact Prior (Realization Only)}: we set $w_i^e{=}1$
during training and retain the contact-aware weights during grasp
realization, resulting in drops of 3.28 and 1.66 pp; and
3) \textit{Contact Prior (Training Only)}: we retain the contact-aware
weights during training but set $w_i^e{=}1$ during grasp realization,
resulting in drops of 1.72 and 1.43 pp, respectively. 
In this setting, we reuse the full model's results for SharpaWave as both settings use $w_i^e{=}1$ during grasp realization.
These results show that incorporating the contact prior during both training and grasp realization improves grasping performance.

\subsection{Real-World Experiments}
To answer \textbf{Q8}, we deploy our pipeline on a KUKA-mounted SharpaWave hand under the same zero-shot setting as Sec.~\ref{sec:unseen_hand}. Object poses are estimated using FoundationPose~\cite{foundationposewen2024} with an Intel RealSense D435 camera. Using the same grasp-generation setup as in simulation,
including the pre-grasp pose initialization, our method generates
100 grasp candidates per object through 150 optimization iterations. The candidates
are considered sequentially: each undergoes arm path planning with the table treated as an obstacle and is executed if a feasible trajectory can be found; otherwise, the next is considered. For execution, the hand starts open at an
approach pose 10~cm behind the target wrist pose along the palm
approach direction, advances to the target wrist pose, and closes to the target joint positions under position control.

We further evaluate a baseline that brings the open hand to a pre-grasp
pose initialized as in Sec.~\ref{sec:recovery} using the estimated object pose,
followed by torque-controlled finger closing. Each joint closes until its torque reaches $0.3\,\mathrm{N\!\cdot\!m}$, after which its torque is maintained at this level. We evaluate both methods on 10 unseen objects with
10 grasp attempts each and report the results in
Table~\ref{tab:real_world_results}. Our method achieves \textbf{86\%} success, compared with \textbf{49\%} for the baseline. Both methods can fail due to residual object-pose estimation errors or early finger contacts during closing that displace the object before the remaining contacts are established. For the baseline, independent joint-level torque control can further disturb established contact. It can also fail by directly using the initialized pre-grasp wrist pose, which may be unsuitable for grasping. In contrast, our method treats the pre-grasp pose only as an initialization and further optimizes the pose during grasp realization.

\begin{table}[!t]
\centering
\caption{Real-world experiment results with the unseen SharpaWave.}
\vspace{-8pt}
\label{tab:real_world_results}
\resizebox{\columnwidth}{!}{
\begin{tabular}{lccccc}
\hline
Object 
& Apple & Tomato Soup Can & Rugby Ball & Spam Can & Ketchup Bottle \\
\hline
Success 
& 6/10$^{*}$\,\textbar\,2/10$^{\dagger}$
& 10/10$^{*}$\,\textbar\,5/10$^{\dagger}$
& 9/10$^{*}$\,\textbar\,5/10$^{\dagger}$
& 8/10$^{*}$\,\textbar\,4/10$^{\dagger}$
& 8/10$^{*}$\,\textbar\,4/10$^{\dagger}$ \\
\hline
Object 
& Toothpaste Box & Chips Can & Coke Can & Salt Box & Scrub Sponge \\
\hline
Success 
& 10/10$^{*}$\,\textbar\,7/10$^{\dagger}$
& 8/10$^{*}$\,\textbar\,6/10$^{\dagger}$
& 8/10$^{*}$\,\textbar\,4/10$^{\dagger}$
& 9/10$^{*}$\,\textbar\,6/10$^{\dagger}$
& 10/10$^{*}$\,\textbar\,6/10$^{\dagger}$ \\
\hline
\end{tabular}
}
\vspace{-6pt}
\begin{flushleft}
\footnotesize
$^{*}$Results for MANGO-Grasp; $^{\dagger}$Results for Baseline.
\end{flushleft}
\vspace{-22pt}
\end{table}

\section{Conclusion}
We presented MANGO-Grasp, an anisotropic interaction framework for cross-embodiment dexterous grasp synthesis. MANGO-Grasp represents objects as geometry-oriented 3D Gaussian primitives and encodes hand keypoints into morpho-kinematic descriptors. It models the interaction between hand keypoints and object primitives as Mahalanobis fields that capture the directional nature of contact. The predicted fields guide grasp realization with one shared optimization setup across hands. Experiments demonstrate strong seen- and unseen-hand performance and effective real-world deployment. Future work will extend this pipeline to closed-loop execution. Contact feedback could be used to refine object-pose estimates by enforcing consistency between contact regions and the object surface. Tactile-visual closed-loop replanning could detect contact loss and trigger visual relocalization for subsequent grasp regeneration. Tactile or joint-torque feedback could detect early contact and trigger compliant finger control. We will also explore pre-grasp pose initialization conditioned
on object geometry and the environment to improve grasp execution feasibility.

\bibliographystyle{IEEEtran}
\bibliography{refs}

\end{document}